\documentclass[letterpaper]{article} % DO NOT CHANGE THIS
\usepackage{aaai2027}
\nocopyright
\usepackage[hyphens]{url}  % DO NOT CHANGE THIS
\usepackage{graphicx} % DO NOT CHANGE THIS
\usepackage{natbib}  % DO NOT CHANGE THIS AND DO NOT ADD ANY OPTIONS TO IT
\usepackage{caption} % DO NOT CHANGE THIS AND DO NOT ADD ANY OPTIONS TO IT
\usepackage{multirow}
\usepackage{algorithm}
\usepackage{algorithmic}
\usepackage{microtype}
\usepackage{graphicx}
\usepackage{subcaption}
\usepackage{booktabs} % for professional tables
\usepackage{array}
\usepackage{pifont}
\usepackage{amsmath}
\usepackage{amssymb}
\usepackage{mathtools}
\usepackage{amsthm}
\usepackage{algorithm}
\usepackage{algorithmic}
\usepackage[table]{xcolor}  % 允许表格单元格上色
\usepackage{booktabs}
\usepackage{tcolorbox}
\usepackage{soul}
\definecolor{bestbg}{RGB}{198,219,239}    % 浅蓝（建议做 best 背景，不要太深）
\definecolor{secondbg}{RGB}{222,235,247}  % 更浅
\definecolor{thirdbg}{RGB}{239,243,255}   % 最浅
\usepackage[capitalize,noabbrev]{cleveref}
\usepackage{newfloat}
\usepackage{listings}
\DeclareCaptionStyle{ruled}{labelfont=normalfont,labelsep=colon,strut=off} % DO NOT CHANGE THIS
\floatstyle{ruled}
\newfloat{listing}{tb}{lst}{}
\floatname{listing}{Listing}
\usepackage{booktabs}
\usepackage{pifont}
\usepackage{booktabs}

\title{Learn to Modify: Self-evolving Multi-agent Refinement Framework for Conversational Text-to-Motion Generation}

\author{
Jianan Li\textsuperscript{\rm1},
Jiawei Zhao\textsuperscript{\rm1},
Yangtao Zhou\textsuperscript{\rm1},
Zhifu Zhao\textsuperscript{\rm2}\corresponding,
Shaoxin Wu\textsuperscript{\rm1},
Yutao Wang\textsuperscript{\rm1},
Dezhen Wang\textsuperscript{\rm3},
Hua Chu\textsuperscript{\rm1},
Fu Li\textsuperscript{\rm2}
}

\affiliations{
\textsuperscript{\rm1}School of Computer Science and Technology, Xidian University, Xi'an, China\\
\textsuperscript{\rm2}School of Artificial Intelligence, Xidian University, Xi'an, China\\

\textsuperscript{\rm3}School of Computer Science and Technology, Tongji University, Shanghai, China\\
lijianan@xidian.edu.cn,
zhao\_jw@stu.xidian.edu.cn,
hchu@mail.xidian.edu.cn,
fuli@mail.xidian.edu.cn
}

\begin{document}

\maketitle
% \begin{figure}
%  \includegraphics[width=\linewidth]{image/conversation.png}
%  \caption{Comparison of single-turn generation and iterative refinement for conversational motion generation. Left: Conventional methods synthesize the full motion sequence from a single prompt, making fine-grained correction difficult and inefficient. Right: Our iterative refinement paradigm supports multi-turn conversational feedback, allowing users to progressively refine motion details while preserving previously established motion context through multi-agent collaboration.}
%   \label{fig:introduction}
% \end{figure}

\begin{abstract}
Despite rapid progress in text-to-motion generation, existing methods are commonly designed for single-shot generation.
When users are dissatisfied with specific motion details, they are often forced to regenerate the entire sequence from scratch, which is computationally inefficient and potentially destroys previously established kinematic context. 
A practical generation system should instead support iterative user feedback and refine only the affected spatio-temporal regions while preserving the remaining motion.
Therefore, we introduce a conversational motion generation paradigm that requires the system to coordinate instruction understanding, motion generation, localized editing, and quality assessment. 
Recent advances in LLM-driven multi-agent systems provide a promising foundation for such role-specialized coordination.
% However, instantiating motion generation with multi-agent refinement remains challenging.  First, personalized modifications must preserve previously accepted motion content and maintain temporal continuity. Second, naive multi-agent collaboration may repeatedly explore similar editing strategies across multi-turn interactions, resulting in redundant exploration.
To tackle these challenges, a \textbf{S}elf-evolving \textbf{M}ulti-\textbf{A}gent \textbf{R}efinemen\textbf{T} Framework  (SMART) is proposed for conversational motion generation. 
Within SMART, to preserve semantic consistency, we design a semantic-guided masked motion editing mechanism, which precisely grounds natural language feedback into spatiotemporal motion regions to enable localized refinements without altering unedited regions. 
To reduce redundant exploration, we propose a memory-based self-evolving strategy that dynamically accumulates motion editing knowledge from interaction history and reuses past successful experience.
Extensive experiments demonstrate that SMART effectively supports multi-turn conversational motion generation and enables efficient motion refinement.
\end{abstract}
\begin{figure}[!t]
    \centering
    \includegraphics[width=\columnwidth]{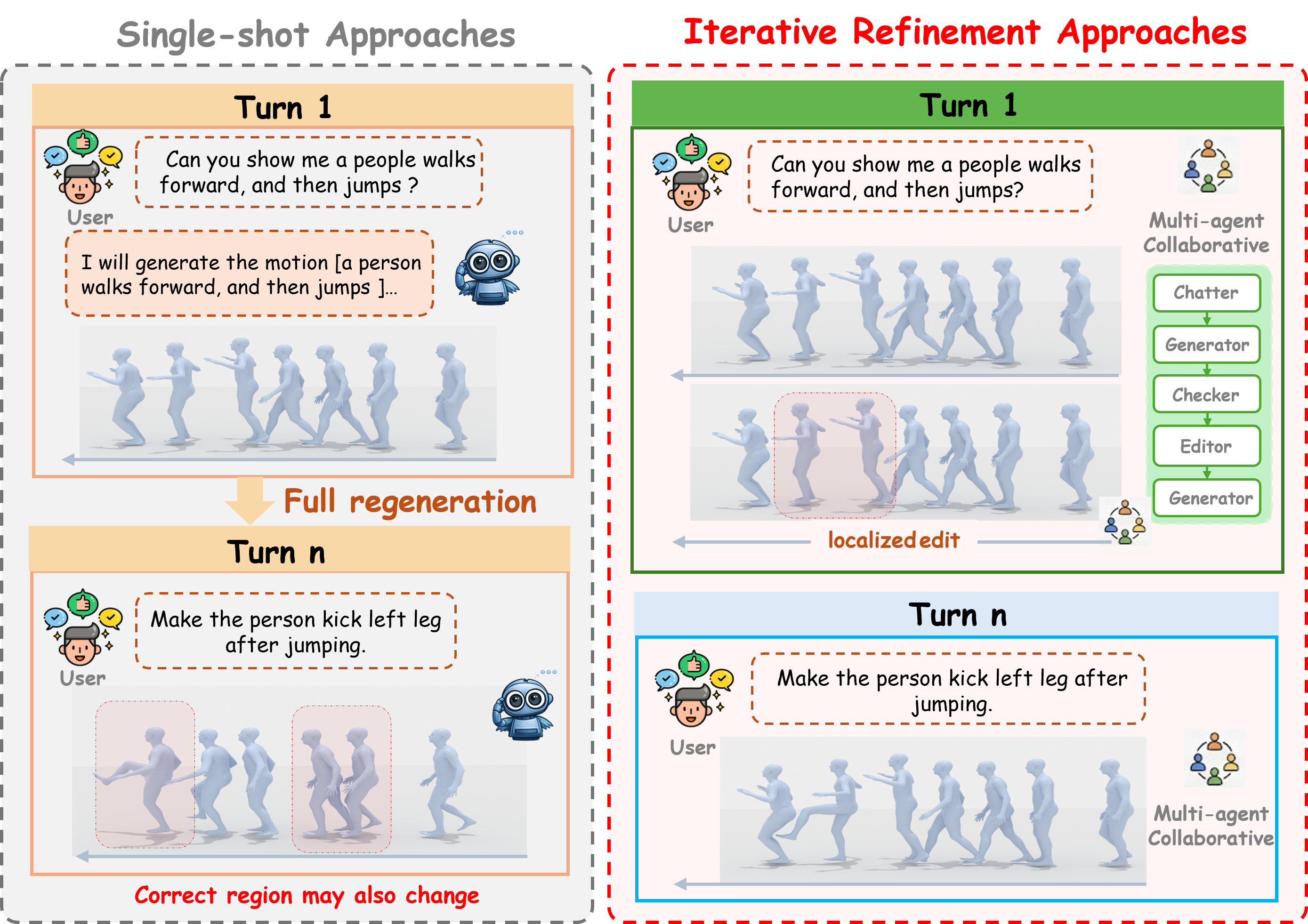}
    \caption{Comparison of single-shot generation and iterative refinement. Conventional methods regenerate the entire motion, hindering fine-grained correction. Our multi-agent framework uses multi-turn feedback to progressively refine motion details while preserving established context.}
    \label{fig:introduction}
\end{figure}
% content/01_intro.tex

\section{Introduction}
Text-to-Motion (T2M) generation~\cite{zhu2023human, chen2024sato, yu2024towards}, which aims to synthesize realistic human motions from natural language descriptions, has emerged as a fundamental task with broad applications in gaming, virtual reality, and robotics~\cite{guo2025tstmotion, tashakori2025flexmotion, zitkovich2023rt}. 
Despite recent progress, T2M generation remains challenging because it requires both accurate language understanding and faithful translation of textual semantics into temporally coherent motion sequences.

Existing T2M methods have explored diverse generation paradigms. 
Autoencoder-based methods~\cite{zhang2023generating,zhong2023attt2m,geng2024text} learn compact latent motion representations for synthesis. GAN-based methods~\cite{ahn2018text2action,xu2023actformer,ahuja2019language2pose} improve motion realism through adversarial training. Autoregressive methods~\cite{jiang2023motiongpt,zhang2023t2m,guo2022tm2t} model motions as discrete token sequences. And diffusion-based methods~\cite{tevet2022human,10.1145/3664647.3681657,10.1145/3581783.3611887} generate motions through iterative denoising. 
Recently, driven by rapid advancements in large language models (LLMs)~\cite{achiam2023gpt,zhao2023survey,10.1145/3746027.3755164}, LLM-based motion generation approaches~\cite{jiang2023motiongpt, wang2024motiongpt, wu2025motion} have made significant strides in generating high-fidelity motions that align with complex textual instructions.

However, most existing approaches are still designed for single-shot generation. 
When users are dissatisfied with specific motion details, they usually have to regenerate the entire sequence from scratch. 
Such a paradigm is computationally inefficient and may discard previously established kinematic context, causing unnecessary changes to motion regions that are already semantically correct. 
This limitation is particularly problematic in real-world creative workflows, where users often refine their intent progressively through feedback rather than specifying all desired details in a single instruction.

These observations motivate a more user-centric setting, namely conversational motion generation, where motions are iteratively refined through multi-turn natural language interaction. 
Unlike single-shot generation, this setting requires the system to coordinate instruction understanding, motion synthesis, localized editing, quality assessment, and long-term interaction memory. 
Recent advances in multi-agent systems~\cite{cardoso2021review, calegari2021logic} provide a promising foundation for such role-specialized coordination. 
Nevertheless, instantiating conversational motion generation with multi-agent refinement remains challenging. 
First, personalized edits usually target only specific body parts or temporal segments, but naive refinement may unintentionally alter already correct motion regions and compromise global semantic consistency. 
Second, without an explicit memory mechanism, agents may repeatedly explore similar editing strategies across interaction turns, leading to redundant reasoning and inefficient refinement.

To address these challenges, we propose \textbf{SMART}, a \textbf{S}elf-evolving \textbf{M}ulti-\textbf{A}gent \textbf{R}efinemen\textbf{T} framework for conversational motion generation. 
SMART organizes multiple role-specialized agents into a closed-loop workflow for instruction rewriting, motion generation, video-text checking, and localized editing.
A semantic-guided masked motion editing mechanism is designed to use a Vision-Language Model (VLM) to ground natural-language feedback into a mask matrix, enabling precise refinement while preserving unaffected motion regions.
To reduce redundant exploration, SMART further incorporates a memory-based self-evolving strategy that
accumulates and updates generation knowledge from interaction history, allowing the system to optimize generation motion over multiple turns.
We further implement an interactive conversational platform to evaluate SMART under realistic multi-turn refinement scenarios. 
Extensive experiments demonstrate that SMART effectively supports conversational motion refinement, preserves unaffected motion regions, and improves refinement efficiency across interaction rounds.
Our contributions are summarized as follows.
The main contributions of this work are summarized as follows:
\begin{itemize}
    \item We formulate \textbf{conversational motion generation} and propose \textbf{SMART}, a self-evolving multi-agent refinement framework that supports iterative text-to-motion refinement through multi-turn natural language interaction.

    \item We design a \textbf{semantic-guided masked motion editing mechanism} that converts structured feedback into spatio-temporal editing masks, enabling localized refinement while preserving unaffected motion regions.

    \item We develop a \textbf{memory-based self-evolving strategy} that accumulates successful failure--repair experiences and reuses them to reduce redundant exploration during multi-turn refinement.
\end{itemize}

% \begin{table*}[t]
% \centering
% \small
% \begin{tabular*}{\textwidth}{
% @{\extracolsep{\fill}}
% lcccccc
% }
% \toprule
% \textbf{Method}
% & \shortstack{\textbf{Multi-turn}\\\textbf{Dialogue}}
% & \shortstack{\textbf{Local}\\\textbf{Editing}}
% & \shortstack{\textbf{Non-edited}\\\textbf{Region Protection}}
% & \shortstack{\textbf{Automatic}\\\textbf{Checking}}
% & \shortstack{\textbf{Cross-turn}\\\textbf{Context}}
% & \shortstack{\textbf{No Extra}\\\textbf{Training}} \\
% \midrule

% Motion-Agent
% & \ding{51}
% & $\triangle$
% & \ding{55}
% & \ding{55}
% & \ding{51}
% & \ding{55} \\

% MotionChain
% & \ding{51}
% & $\triangle$
% & \ding{55}
% & \ding{55}
% & \ding{51}
% & \ding{55} \\

% MotionGPT-2
% & \ding{55}
% & \ding{55}
% & \ding{55}
% & \ding{55}
% & \ding{55}
% & \ding{55} \\

% MotionFix
% & \ding{55}
% & $\triangle$
% & $\triangle$
% & \ding{55}
% & \ding{55}
% & \ding{55} \\

% \textbf{MIRO}
% & \ding{51}
% & \ding{51}
% & \ding{51}
% & \ding{51}
% & \ding{51}
% & \ding{51} \\

% \bottomrule
% \end{tabular*}

% \caption{Capability comparison with representative motion generation
% and editing methods. \ding{51} denotes explicit support,
% $\triangle$ denotes partial or implicit support, and \ding{55} denotes
% that the capability is not supported or reported.}
% \label{tab:capability_comparison}
% \end{table*}

\section{Related Work}
\label{sec:Related Work}
%-------------------------------------------------------------------------
\subsection{Text-to-Motion Generation}
Existing text-to-motion (T2M) methods can be broadly categorized into autoencoder-based, GAN-based, autoregressive, and diffusion-based approaches. Autoencoder-based methods \cite{zhang2023generating,zhong2023attt2m,geng2024text} learn compact latent representations to capture motion structures, while GAN-based methods \cite{ahn2018text2action,xu2023actformer,ahuja2019language2pose} employ adversarial learning to improve motion realism and diversity. Autoregressive methods generate discrete motion tokens sequentially. For example, MoMask \cite{guo2024momask} adopts masked token prediction with residual refinement to improve generation efficiency, whereas MotionChain \cite{jiang2024motionchain} and Motion-Agent \cite{wu2025motion} incorporate language-based planning to support more complex motion tasks. Diffusion-based methods iteratively denoise motion sequences conditioned on textual descriptions. MotionDiffuse \cite{zhang2024motiondiffuse} introduces text-conditioned diffusion with temporal modeling, MDM \cite{tevet2022human} directly predicts clean motion samples with geometric losses, and ReMoDiffuse \cite{zhang2023remodiffuse} enhances text-motion alignment through retrieval-augmented generation. 
\begin{figure*}[!t]
\centering

\includegraphics[width=0.95\linewidth]{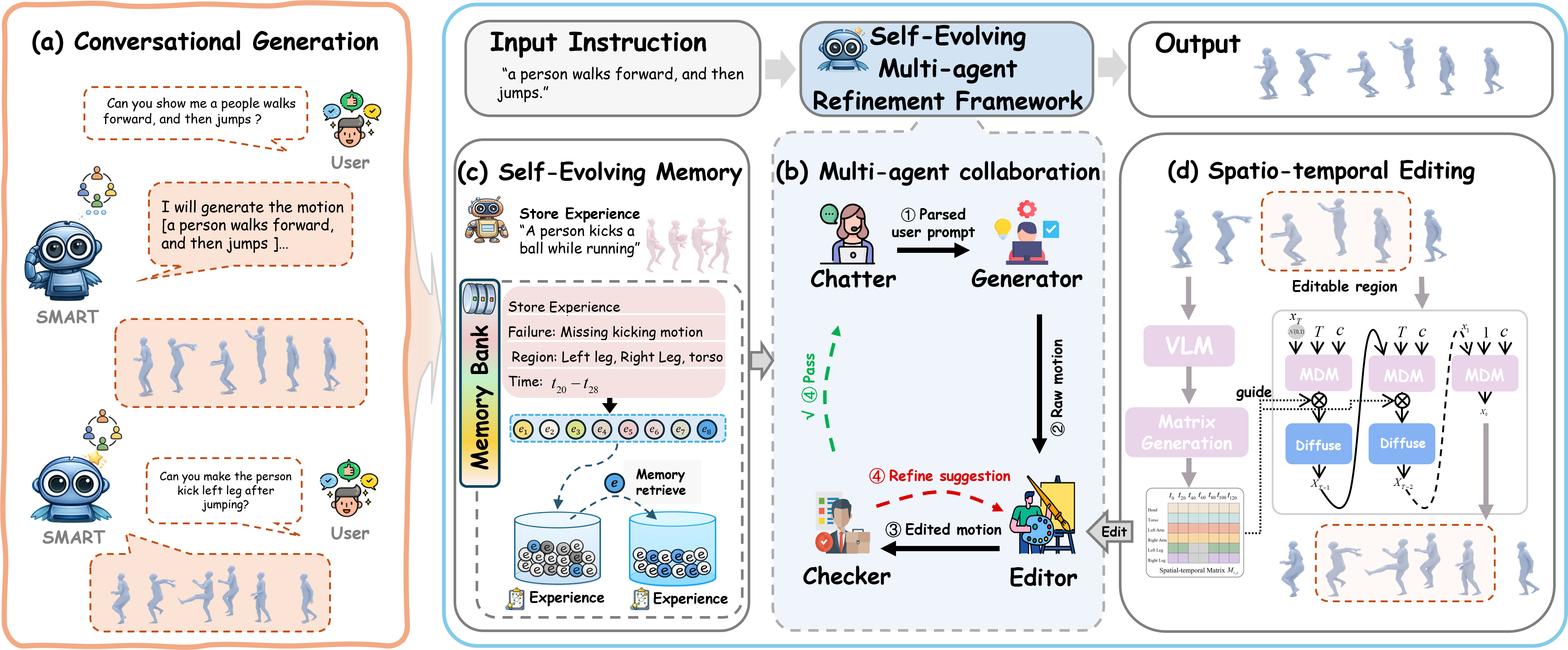}
\caption{
Overview of SMART.
(a) Iterative motion refinement through conversational motion generation. 
(b) Multi-agent collaboration among the Chatter, Generator, Editor, and Checker for instruction alignment, motion synthesis, localized editing, and semantic verification.
(c) Self-evolving strategy that accumulates generation experiences for motion refinement. 
(d) Semantic-guided motion editing that localizes user feedback with a Vision-Language Model (VLM) and refines selected motion regions while preserving unmasked content. 
}
\label{fig:method}
\end{figure*}
%-------------------------------------------------------------------------

\subsection{LLM-based motion generation}
Large language models (LLMs) have demonstrated strong capabilities in language understanding, task decomposition, tool invocation, and self-correction \cite{chang2024survey,ruan2023tptu,wu2023embodied,li2023tradinggpt,wang2025g,wang2024videoagent}. These capabilities have motivated agent-based generative systems that coordinate specialized tools for complex generation and editing tasks \cite{durante2024agent,masterman2024landscape}. For instance, GenArtist \cite{wang2024genartist} employs a multimodal agent to select and coordinate visual tools, demonstrating the potential of agents to unify generation and editing.
In motion generation, MotionGPT \cite{jiang2023motiongpt} represents motions as discrete tokens and unifies text and motion modeling within a language-modeling framework. MotionGPT-2 \cite{wang2024motiongpt} further extends this paradigm to motion generation, understanding, and editing. 
We summarize the capabilities of representative motion generation and editing methods in Table~\ref{tab:capability_comparison}, comparing them across multi-turn dialogue, motion editing, automatic checking, cross-turn Context, and no extra training requirements.
% 继续引出其他的AI Agent的文章，介绍，引出自己的方法。

\begin{table}[hb]
\centering

\small
\setlength{\tabcolsep}{2.5pt}
\resizebox{\columnwidth}{!}{
\begin{tabular}{lccccc}
\toprule
\textbf{Method}
& \shortstack{\textbf{Multi-turn}\\\textbf{Dialogue}}
& \shortstack{\textbf{Motion}\\\textbf{Editing}}
& \shortstack{\textbf{Automatic}\\\textbf{Checking}}
& \shortstack{\textbf{Cross-turn}\\\textbf{Context}}
& \shortstack{\textbf{No Extra}\\\textbf{Training}} \\
\midrule

Motion-Agent
& \ding{51}
& \ding{51}
& \ding{55}
& \ding{51}
& \ding{55} \\

MotionChain
& \ding{51}
& \ding{51}
& \ding{55}
& \ding{51}
& \ding{55} \\

MotionGPT-2
& \ding{55}
& \ding{51}
& \ding{55}
& \ding{55}
& \ding{55} \\

MotionFix
& \ding{55}
& \ding{51}
& \ding{55}
& \ding{55}
& \ding{55} \\

\textbf{SMART}
& \ding{51}
& \ding{51}
& \ding{51}
& \ding{51}
& \ding{51} \\

\bottomrule
\end{tabular}
}
\caption{Capability comparison with representative motion generation and editing methods. \ding{51} and \ding{55} denote supported and unsupported capabilities, respectively.}
\label{tab:capability_comparison}
\end{table}

\section{Problem Formulation}
\label{sec: Problem Formulation}
For multi-turn interactive human motion generation, the system continuously updates a motion according to a sequence of user instructions while maintaining consistency with previously accepted interaction outcomes.

Let $\mathcal{U}$ and $\mathbf{X}$ denote the user instructions and human motions, respectively. A motion sequence is represented as ${\bf{x}} \in {\mathbb{R}^{T \times D}}$, where $T$ denotes the sequence length and $D$ is the dimensionality of the motion representation at each frame. 
And let $u^{(0)} \in \mathcal{U}$ denote the initial motion description and $u^{(k)} \in \mathcal{U}$ denote the user instruction at the $k$-th interaction round, where $k \geq 1$. The initial motion is defined as $\mathbf{X}^{(0)}$. 
After each user-level interaction round, the system maintains an interaction history defined as
\begin{equation}
\mathcal{H}^{(k)}
=
\left\{
\left(
u^{(i)},
\mathbf{X}^{(i)},
r^{(i)}
\right)
\right\}_{i=0}^{k},
\label{eq:interaction_history}
\end{equation}
where $\mathbf{m}^{(i)}$ is the accepted motion after the $i$-th interaction round, and $r^{(i)}$ denotes the corresponding interaction record, such as \textit{\textbf{the verification outcome, refinement summary, or user confirmation}}. 
For a subsequent user instruction $u^{(k)}$, the system updates the motion by jointly considering the previously accepted motion and the historical interaction context:
\begin{equation}
\mathbf{X}^{(k)}
=
\mathcal{P}
\left(
\mathbf{X}^{(k-1)},
u^{(k)},
\mathcal{H}^{(k-1)}
\right),
\qquad k \geq 1,
\label{eq:user_level_interaction}
\end{equation}
where $\mathcal{P}$ denotes the interactive motion generation process. Additionally, user instructions in each turn are resolved through multiple internal multi-agent refinement cycles.
Specifically, let ${\mathbf{x}}^{(k)}_{\ell}$ denote the candidate motion obtained at the $\ell$-th internal refinement cycle for the $k$-th user instruction, where $\ell = 0,\ldots,L_{\max}$ and $L_{\max}$ is the maximum number of refinement cycles. 
At each internal cycle, the multi-agent system takes the current candidate motion and the available experience memory as input to produce an updated candidate together with verification feedback:
\begin{equation}
\left(
\tilde{\mathbf{X}}^{(k)}_{\ell},
r^{(k)}_{\ell},
\mathcal{B}^{(k)}_{\ell}
\right)
=
\Phi_{\mathrm{SMART}}
\left(
\tilde{\mathbf{X}}^{(k)}_{\ell-1},
\mathcal{B}^{(k)}_{\ell-1}
\right),
\label{eq:agent_refinement}
\end{equation}
where $\Phi_{\mathrm{SMART}}$ denotes one internal multi-agent refinement cycle, $r^{(k)}_{\ell}$ is the verification feedback for the current candidate, and $\mathcal{B}^{(k)}_{\ell}$ represents the available experience memory after the $\ell$-th refinement cycle. This formulation allows SMART to iteratively revise a candidate motion according to intermediate feedback, rather than relying on a single-pass generation outcome.

\section{Self-evolving Multi-agent Refinement Framework}

An overview of SMART is illustrated in Fig.~\ref{fig:method}. 
SMART formulates conversational motion generation as a multi-agent self-evolving refinement process. 
The framework operates at two levels. 
At the outer level, the user progressively refines an initially generated motion through a sequence of natural-language instructions, as shown in Fig.~\ref{fig:method}(a). 
At the inner level, each user instruction is resolved through agent-driven refinement cycles within SMART. It adopts a role-specialized directed workflow with explicit motion states and verification feedback. 
Moreover, SMART introduces a memory-based self-evolving mechanism that records successful refinement experiences and reuses them in later interaction rounds. 
In this way, the framework not only supports closed-loop motion refinement within each turn but also improves refinement efficiency across turns through accumulated historical experience.

\subsection{Multi-agent Workflow}

SMART decomposes the overall process into several role-specialized agents with explicit interfaces.
Specifically, 
Chatter and Generator initialize a motion proposal, whereas Editor and Checker iteratively refine it through a closed-loop feedback mechanism.
 
\textbf{Chatter Agent} 
To resolve semantic ambiguity, non-standard, or context-
dependent expression, the Chatter Agent rewrites raw user inputs into standardized and motion-oriented descriptions before motion synthesis.

Given the current user instruction $u^{(k)}$ and the dialogue history $\mathcal{H}^{(k-1)}$, the LLM-based prompt synthesizer $\mathcal{P}_{\mathrm{chat}}$ rewrites the raw input into a standardized motion description:
\begin{equation}
C^{(k)}
=
\mathcal{P}_{\mathrm{chat}}
\left(
\mathcal{H}^{(k-1)}, u^{(k)}
\right).
\label{eq:chatter_rewrite}
\end{equation}

The prompt used by the LLM-based prompt synthesizer $\mathcal{P}_{\mathrm{chat}}$ is defined as follows:
\begin{tcolorbox}[colback=gray!10, colframe=gray!50, boxrule=0.5pt, arc=2pt]
\small
\textit{
Act as a motion description rewriting agent. 
Given the dialogue \textcolor{magenta}{<History>} and the current \textcolor{blue}{<User Input>}, rewrite the input into a standardized motion description. 
The rewritten description should preserve the original user intent while resolving ambiguous, non-standard, or colloquial expressions. 
Output only the rewritten \textcolor{red}{<MotionDescription>} without any explanation.
}
\end{tcolorbox}

\textbf{Generator Agent}
Given the standardized motion description $C^{(k)}$ produced by Chatter, the Generator Agent aims to synthesize a corresponding human motion sequence. 
In our implementation, we use an existing text-to-motion model as an external tool for generating human motions.
Notably, SMART does not rely on a specific motion generator.
Formally, conditioned on the rewritten motion description $C^{(k)}$, Generator samples a raw motion proposal $\mathbf{X}_{\mathrm{raw}}^{(k)} \in \mathbb{R}^{L \times D}$ from the existing text-to-motion model, where $L$ is the number of frames, and $D$ is the dimensionality of the motion feature at each frame. 
For the first interaction round, $\mathbf{X}_{\mathrm{raw}}^{(0)}$ serves as the initial motion, while in subsequent rounds it provides a raw proposal for verification and refinement.

\textbf{Checker Agent} 
The Checker Agent verifies whether a candidate motion is semantically consistent with the standardized motion description and determines whether it can be accepted. 
Since user instructions often involve open-vocabulary and compositional semantics, feature-level distances or predefined action classifiers are insufficient for reliable verification. 

Checker first renders the candidate motion into a video $V^{(k)}_{\ell}$ by converting the generated motion representation into 3D joint coordinates. 
Then, the rendered video $V^{(k)}_{\ell}$ and the standardized motion description $C^{(k)}$ are fed into a VLM-based evaluation function $\Psi(\cdot)$ to produce a structured feedback report $F^{(k)}_{\ell}$. 
In our implementation, we use Qwen-VL~\cite{Qwen-VL} as the visual semantic evaluator, and the detailed prompt design is provided in the Appendix.
% This allows Checker to detect fine-grained inconsistencies, such as incorrect body parts, missing actions, wrong temporal order, or mismatched motion styles.
Beyond a binary accept-or-reject decision, our feedback report is decomposed into three components:
\begin{equation}
F^{(k)}_{\ell}
=
\left(
O^{(k)}_{\ell},
\mathcal{T}^{(k)}_{\ell},
E^{(k)}_{\ell}
\right),
\label{eq:checker_feedback_decomposition}
\end{equation}
where $O^{(k)}_{\ell}$ denotes the body-part regions that require refinement, $\mathcal{T}^{(k)}_{\ell}$ denotes the temporal intervals where the inconsistency occurs, and $E^{(k)}_{\ell}$ denotes the corresponding editing instruction.
If the candidate motion is judged to be satisfactory, Checker outputs an empty feedback report, $F^{(k)}_{\ell}= \varnothing$, and the current interaction can be stored as a successful experience. 
Otherwise, $F^{(k)}_{\ell}$ is passed to the Editor agent for localized refinement.

\textbf{Editor Agent}  
The Editor Agent converts the structured feedback into a localized refinement operation. 
Instead of regenerating the entire motion sequence, the Editor performs targeted motion editing based on the detected error regions.
Specifically, when the feedback report is non-empty, i.e.,
$F^{(k)}_{\ell}\neq \varnothing$, Editor takes the current candidate motion $\mathbf{X}^{(k)}_{\ell}$
% , the standardized motion description $C^{(k)}$, 
and the feedback report $F^{(k)}_{\ell}$ as inputs, and produces a refined motion candidate for the next refinement cycle:
\begin{equation}
\hat{\mathbf{X}}^{(k)}_{\ell+1}
=
\mathcal{E}_{\mathrm{edit}}
\left(
\mathbf{X}^{(k)}_{\ell}
,
% C^{(k)},
F^{(k)}_{\ell}
\right),
\label{eq:editor_refinement}
\end{equation}
where $\mathcal{E}_{\mathrm{edit}}(\cdot)$ denotes the semantic-guided masked motion editing process.
After editing, the refined candidate $\hat{\mathbf{X}}^{(k)}_{\ell+1}$ is sent back to Checker for another round of video-text semantic verification. 
This Editor--Checker loop continues until the candidate motion is accepted or the maximum refinement budget is reached. 
This editing strategy will be detailed in the next section. 

\subsection{Semantic-guided Masked Motion Editing Mechanism}
\label{section:spatio-temporal masked}

Instead of global regeneration, we introduce a semantic-guided masked motion editing mechanism that updates only the regions indicated by Checker while preserving the remaining motion content.
Given the candidate motion $\mathbf{X}^{(k)}_{\ell}$ and a non-empty feedback report 
$F^{(k)}_{\ell}=(O^{(k)}_{\ell},\mathcal{T}^{(k)}_{\ell}, E^{(k)}_{\ell})$ from Checker, the editing module first parses the feedback into executable spatio-temporal constraints. 
The body-part feedback $O^{(k)}_{\ell}$ is selected from a predefined six-part body partition:
\begin{equation}
\mathcal{O}
=
\{o_0,o_1,o_2,o_3,o_4,o_5\},
\end{equation}
where the six regions correspond to the head, torso, left arm, right arm, left leg, and right leg, respectively. 
Each selected body part is mapped to its associated skeletal joints and then to the corresponding motion feature channels according to the adopted motion representation. 
The temporal feedback $\mathcal{T}^{(k)}_{\ell}=\{\tau_j\}_{j=1}^{n}$ specifies the erroneous intervals, where each interval is denoted as $\ tau_j = (t_j^s,t_j^e)$. 
The editing instruction $E^{(k)}_{\ell}$ provides a short semantic correction, such as ``make the right arm wave'' or ``replace the sitting action with turning around'', and is used to construct the local editing condition.

Based on the parsed body-part and temporal constraints, we construct a binary spatio-temporal mask:
\begin{equation}
\mathbf{M}^{(k)}_{\ell}
\in
\{0,1\}^{L\times D},
\end{equation}
where $L$ denotes the number of frames and $D$ denotes the dimensionality of the motion feature. 
The mask is defined as
\begin{equation}
\mathbf{M}^{(k)}_{\ell}(l,d)
=
\begin{cases}
1, & \text{if } (l,d) \text{ belongs to the editable region},\\
0, & \text{otherwise},
\end{cases}
\label{eq:editing_mask}
\end{equation}
where $l$ indexes the frame and $d$ indexes the motion feature channel. 
To avoid abrupt transitions between edited and preserved regions, we apply temporal smoothing to the binary mask:
\begin{equation}
\hat{\mathbf{M}}^{(k)}_{\ell}
=
\mathbf{M}^{(k)}_{\ell}
*
\mathcal{K},
\label{eq:mask_smoothing}
\end{equation}
where $\mathcal{K}$ is a one-dimensional Gaussian kernel applied along the temporal axis. 
The resulting soft mask $\hat{\mathbf{M}}^{(k)}_{\ell}\in[0,1]^{L\times D}$ enables smooth blending around editing boundaries.

Next, the smoothed mask matrix $\hat{\mathbf{M}}^{(k)}_{\ell}$ combined with the editing instruction $E^{(k)}_{\ell}$  to perform motion editing under the mask guidance:
\begin{equation}
\mathbf{X}^{(k)}_{\ell+1} \sim P_G(\cdot \mid\mathbf{X}^{(k)}_{\ell} \odot (1-\hat{\mathbf{M}}^{(k)}_{\ell}),E_{\ell}^{(k)}),
\end{equation}
where $\odot$ denotes element-wise multiplication, $P_G$ is instantiated by t2m model in our implementation. 
The refined motion $\mathbf{X}^{(k)}_{\ell+1}$ is sent back to Checker for video-text semantic verification. 
If accepted, it is committed as the output of the current interaction round; otherwise, the Editor--Checker loop continues until the maximum refinement budget is reached.

\subsection{Memory-based Self-Evolving Strategy}
\label{section:Self-Evolving Strategy}

We introduce a memory-based self-evolving strategy that stores successful failure-repair experiences in a memory bank and retrieves relevant cases for later editing rounds. 
Different from online model updating, SMART evolves at the experience level by accumulating reusable motion editing knowledge from historical interactions.

\noindent\textbf{Memory bank construction}
In the memory bank $\mathcal{B}$, each entry corresponds to a case where the Editor corrects the inconsistency detected by the Checker and then accepts it in subsequent verification. 
Formally, the $j$-th experience entry $e_j \in \mathcal{B}$ is represented as
\begin{equation}
e_j
=
\left(
\Lambda_j,
E_j,
O_j,
\mathcal{T}_j
\right),
\label{eq:memory_entry}
\end{equation}
where $\Lambda_j$ denotes the semantic keyword set of the refinement case, $E_j$ denotes the editing instruction, $O_j$ denotes the involved body-part regions, and $\mathcal{T}_j$ denotes the temporal region of the correction.

\noindent\textbf{Memory update}
The memory update is triggered when a previously failed candidate is successfully refined and accepted by Checker. 
Suppose Checker produces a non-empty feedback report 
$F^{(k)}_{\ell}=(O^{(k)}_{\ell},\mathcal{T}^{(k)}_{\ell},E^{(k)}_{\ell})$ 
at the $\ell$-th refinement cycle, and the edited motion is later accepted. 
SMART extracts a semantic keyword set from the standardized description and the feedback:
\begin{equation}
\Lambda^{(k)}_{\ell}
=
\operatorname{Extract}
\left(
C^{(k)},
F^{(k)}_{\ell}
\right),
\label{eq:keyword_extraction}
\end{equation}
where $\operatorname{Extract}(\cdot)$ denotes a prompt-based LLM keyword extraction function that captures action types, body parts, temporal cues, and motion attributes. 
The successful refinement experience is then written into the memory bank:
\begin{equation}
\mathcal{B}
\leftarrow
\mathcal{B}
\cup
\left\{
\left(
\Lambda^{(k)}_{\ell},
E^{(k)}_{\ell},
O^{(k)}_{\ell},
\mathcal{T}^{(k)}_{\ell}
\right)
\right\}.
\label{eq:memory_update}
\end{equation}

\noindent\textbf{Memory retrieval}
When Checker produces a non-empty feedback report in a later refinement cycle, SMART retrieves relevant historical experiences to assist the current editing process. 
Given the query keyword set $\Lambda^{(k)}_{\ell}$ and a memory entry 
$e_j=(\Lambda_j, E_j, O_j,\mathcal{T}_j)$, SMART retrieves the most relevant experience by maximizing the Jaccard overlap between keyword sets:
\begin{equation}
e^{*}
=
\arg\max_{e_j\in\mathcal{B}}
\frac{
\left|
\Lambda^{(k)}_{\ell}
\cap
\Lambda_j
\right|
}{
\left|
\Lambda^{(k)}_{\ell}
\cup
\Lambda_j
\right|
},
\quad
\text{s.t. }
O^{(k)}_{\ell}
\cap
O_j
\neq
\emptyset.
\label{eq:memory_retrieval}
\end{equation}
Here, $O^{(k)}_{\ell}$ denotes the body-part regions identified in the current feedback, while $O_j$ denotes the body-part regions stored in the historical experience. 
This constraint filters out experiences that are semantically similar but spatially unrelated, ensuring that the retrieved memory provides editing cues for relevant body regions.

\noindent\textbf{Memory-guided feedback enhancement}
The retrieved experience $e^*=(\Lambda^*,E^*,O^*,\mathcal{T}^*)$ serves as a memory prior for the current feedback 
$F^{(k)}_{\ell}=(O^{(k)}_{\ell},\mathcal{T}^{(k)}_{\ell},E^{(k)}_{\ell})$. 
Rather than directly replacing the current feedback, SMART integrates them to construct memory-enhanced feedback:
\begin{equation}
\bar{F}^{(k)}_{\ell}
=
\operatorname{Enhance}
\left(
F^{(k)}_{\ell},
e^*
\right)
=
\left(
\bar{O}^{(k)}_{\ell},
\bar{\mathcal{T}}^{(k)}_{\ell},
\bar{E}^{(k)}_{\ell}
\right),
\label{eq:memory_enhanced_feedback}
\end{equation}
where $\bar{O}^{(k)}_{\ell}$ expands the current body-part regions with the retrieved experience, 
$\bar{\mathcal{T}}^{(k)}_{\ell}$ merges the current and historical temporal intervals, and 
$\bar{E}^{(k)}_{\ell}$ fuses the current and historical editing instructions through prompt-based text merging. 
Finally, the Editor performs masked motion editing with the enhanced feedback.

\begin{table*}[htbp]
\centering
\setlength{\tabcolsep}{3.2pt}
\small
\resizebox{\textwidth}{!}{
\begin{tabular}{llcccccc}
\toprule
\textbf{Datasets}
& \textbf{Methods}
& \textbf{Top-1$\uparrow$}
& \textbf{Top-2$\uparrow$}
& \textbf{Top-3$\uparrow$}
& \textbf{FID$\downarrow$}
& \textbf{MM Dist$\downarrow$}
& \textbf{Diversity$\rightarrow$} \\
\midrule
% ==================== HumanML3D ====================
\multirow{9}{*}{HumanML3D}

& Real Motion
& 0.511$\pm$.003
& 0.703$\pm$.003
& 0.797$\pm$.002
& 0.002$\pm$.000
& 2.974$\pm$.008
& 9.503$\pm$.065 \\

\cmidrule(lr){2-8}

& MDM \cite{tevet2022human}
& 0.320$\pm$.005
& 0.498$\pm$.004
& 0.611$\pm$.007
& 0.544$\pm$.044
& 5.566$\pm$.027
& 9.559$\pm$.086 \\

& MLD \cite{chen2023executing}
& 0.481$\pm$.003
& 0.673$\pm$.003
& 0.772$\pm$.002
& 0.473$\pm$.013
& 3.196$\pm$.010
& 9.724$\pm$.082 \\

& MotionDiffuse \cite{zhang2024motiondiffuse}
& 0.491$\pm$.001
& 0.681$\pm$.001
& 0.782$\pm$.001
& 0.630$\pm$.001
& 3.113$\pm$.001
& 9.410$\pm$.049 \\

& T2M-GPT \cite{zhang2023t2m}
& 0.491$\pm$.003
& 0.680$\pm$.003
& 0.775$\pm$.002
& 0.116$\pm$.004
& 3.118$\pm$.011
& 9.761$\pm$.081 \\

& MoMask \cite{guo2024momask}
& 0.521$\pm$.002
& 0.713$\pm$.002
& 0.807$\pm$.002
& 0.045$\pm$.002
& 2.958$\pm$.008
& -- \\

& UniTMGE \cite{wang2025unitmge}
& 0.505$\pm$.003
& 0.687$\pm$.003
& 0.781$\pm$.002
& 0.339$\pm$.009
& 3.087$\pm$.008
& 9.521$\pm$.053 \\

& EnergyMoGen \cite{zhang2025energymogen}
& 0.523$\pm$.003
& 0.715$\pm$.002
& 0.815$\pm$.002
& 0.188$\pm$.006
& 2.915$\pm$.007
& 9.488$\pm$.099 \\

& Motion-Agent \cite{wu2025motion}
& 0.515$\pm$.004
& --
& 0.801$\pm$.004
& 0.230$\pm$.009
& 2.967$\pm$.020
& 9.908$\pm$.102 \\

& SPAM \cite{sun2026coma}
& \textbf{0.526$\pm$.003}
& 0.713$\pm$.003
& 0.804$\pm$.002
& 0.108$\pm$.004
& 2.939$\pm$.008
& -- \\

& MOGO  \cite{fu2026mogo}
& 0.515$\pm$.003
& 0.709$\pm$.003
& 0.801$\pm$.003
& \textbf{0.064$\pm$.002}
& 2.951$\pm$.008
& -- \\

\cmidrule(lr){2-8}

& \textbf{SMART (Ours)}
& 0.525$\pm$.003
& \textbf{0.723$\pm$.003}
& \textbf{0.827$\pm$.002}
& 0.075$\pm$.003
& \textbf{2.901$\pm$.016}
& 9.277$\pm$.103 \\

\midrule

% ====================== KIT-ML ======================
\multirow{9}{*}{KIT-ML}

& Real Motion
& 0.424$\pm$.005
& 0.649$\pm$.006
& 0.779$\pm$.006
& 0.031$\pm$.004
& 2.788$\pm$.012
& 11.080$\pm$.097 \\

\cmidrule(lr){2-8}

& MDM \cite{tevet2022human}
& 0.164$\pm$.004
& 0.291$\pm$.004
& 0.396$\pm$.004
& 0.497$\pm$.021
& 9.191$\pm$.022
& 10.847$\pm$.109 \\

& MLD \cite{chen2023executing}
& 0.390$\pm$.008
& 0.609$\pm$.008
& 0.734$\pm$.007
& 0.404$\pm$.027
& 3.204$\pm$.027
& 10.800$\pm$.105 \\

& MotionDiffuse \cite{zhang2024motiondiffuse}
& 0.417$\pm$.004
& 0.621$\pm$.004
& 0.739$\pm$.004
& 1.954$\pm$.062
& 2.958$\pm$.005
& 11.100$\pm$.143 \\

& T2M-GPT \cite{zhang2023t2m}
& 0.416$\pm$.006
& 0.627$\pm$.006
& 0.745$\pm$.006
& 0.514$\pm$.029
& 3.007$\pm$.023
& 10.921$\pm$.108 \\

& MoMask \cite{guo2024momask}
& 0.433$\pm$.007
& 0.656$\pm$.005
& 0.781$\pm$.005
& 0.204$\pm$.011
& 2.779$\pm$.022
& 10.852$\pm$.109 \\

& EnergyMoGen \cite{zhang2025energymogen}
& 0.436$\pm$.006
& 0.651$\pm$.006
& 0.772$\pm$.006
& 0.495$\pm$.020
& 2.861$\pm$.020
& 11.060$\pm$.101 \\

& MOGO \cite{fu2026mogo}
& 0.447$\pm$.023
& 0.668$\pm$.016
& \textbf{0.801$\pm$.007}
& \textbf{0.191$\pm$.009}
& 2.849$\pm$.007
& -- \\

\cmidrule(lr){2-8}

& \textbf{SMART (Ours)}
& \textbf{0.450$\pm$.014}
& \textbf{0.670$\pm$.018}
& 0.784$\pm$.007
& 0.255$\pm$.033
& \textbf{2.587$\pm$.015}
& 10.898$\pm$.399 \\

\bottomrule
\end{tabular}}

\caption{Comparison results on the HumanML3D and KIT-ML test sets.
The best results among generative methods are highlighted in bold.
Higher is better for R-Precision, while lower is better for FID and
MM Dist. Diversity values closer to Real Motion are better.}
\label{tab:motion_comparison}
\end{table*}

\section{Experiments}

% We evaluate SMART on HumanML3D and KIT-ML using both standard and refinement-oriented metrics, including comparisons with state-of-the-art methods, component ablations, multi-turn refinement evaluation, human and VLM assessments, and qualitative analysis. Additional implementation details are provided in the Appendix.
\subsection{Experimental Setup}
To comprehensively evaluate the performance and generalizability of our interactive motion generation framework, 
% we conduct experiments on the benchmark dataset: HumanML3D~\cite{guo2022generating} and KIT-ML~\cite{plappert2016kit}. This dataset spans a range of motion types, annotation styles, and semantic complexity, allowing us to assess our system’s capabilities across both standard and challenging scenarios.

We conduct experiments on two benchmark datasets: HumanML3D~\cite{guo2022generating} and KIT-ML~\cite{plappert2016kit}. These datasets cover diverse motion types, annotation styles, and levels of semantic complexity, enabling a comprehensive evaluation across different scenarios. 
Note that, for the comparison experiments, we instantiate SMART with SALAD as the motion generator on both datasets. 
For ablation studies, we use an MDM-based generator on HumanML3D unless otherwise specified.

\subsubsection{Evaluation metrics}
Our evaluation consists of two parts.
First, we adopt standard text-to-motion metrics to measure the overall quality of generated motions. 
Since standard T2M metrics mainly evaluate single-shot generation, we further introduce refinement-oriented metrics carefully designed for our conversational motion generation setting to assess the effectiveness of iterative motion editing.
\textbf{(1) Standard Text-to-Motion Metrics.}
We adopt standard quantitative metrics, including R-Precision,
FID, MM Dist and Diversity, following previous
works~\cite{guo2022generating} on the HumanML3D and KIT-ML test set.
These metrics jointly evaluate semantic alignment, distributional fidelity, and diversity of the generated motions.
% \begin{itemize}
%      \item \textbf{R-Precision} evaluates text-motion semantic alignment by ranking candidate text descriptions according to their distances from a generated motion in the shared embedding space. It reports the proportion of paired text descriptions retrieved within the top-$K$ candidates.
%     \item \textbf{Fréchet Inception Distance (FID)} measures the distributional similarity between generated and real motions by computing the Fréchet distance between their feature distributions. A lower FID indicates that the generated motions are closer to real motions.
%     \item \textbf{Multimodal Distance (MM Dist.)} measures text-motion semantic alignment by computing the average distance between paired text and generated-motion embeddings in the shared feature space. A lower value indicates better semantic consistency.
%     \item \textbf{Diversity} measures the overall variation of the generated motion distribution by computing the average feature distance between randomly sampled motion pairs. A value closer to that of real motions indicates more realistic distribution-level diversity.
%     \item \textbf{Multimodality} measures the conditional diversity of motions generated from the same text description by averaging the feature distances among multiple samples conditioned on an identical input. A higher value indicates greater intra-text diversity.
% \end{itemize}
\textbf{(2) Conversational Motion Evaluation Metrics.}
% We evaluate Text-Motion Similarity (TMS), Accepted Rate (AR), and Local Preservation Ability (LPA).
% TMS measures the semantic consistency between the generated motion and the target instruction by computing the cosine similarity between motion and text embeddings. AR calculates the proportion of editing operations accepted by the Checker among all refinement attempts, reflecting the effectiveness of local editing. LPA measures whether
% unmodified regions are preserved after refinement by evaluating the similarity between preserved regions before and after editing.
We further evaluate Text-Motion Similarity (TMS), Refinement Success Rate (RSR), and Local Preservation Ability (LPA). TMS measures the semantic consistency between the generated
motion and the target instruction using cosine similarity between text and motion embeddings. RSR reports the proportion of refinement cases accepted by the Checker within the allowed
refinement budget, reflecting the effectiveness of the closed-loop refinement process. LPA measures the similarity of unedited motion regions before and after refinement.

\subsection{Comparison with State-of-the-Art Methods}
\label{sec: Comparison with Motion Generation Methods}
Table~\ref{tab:motion_comparison} compares SMART with representative methods on HumanML3D and KIT-ML. On HumanML3D, SMART (MDM) achieves the best Top-2 and Top-3 R-Precision of 0.723 and 0.827, respectively, while obtaining the lowest MM Dist of 2.901.
% among the compared generative methods. 
On KIT-ML, SMART achieves the best Top-1 and Top-2 scores of 0.450 and 0.670, and reduces MM Dist to 2.587.
% SMART also improves the R-Precision of both MDM and MoMask, and SMART (MoMask) obtains the lowest MM Dist. 
These results demonstrate that SMART consistently improves text-motion semantic alignment while maintaining competitive motion quality.
% , although its effect on distributional metrics varies with the dataset and generator.

% Single-turn metrics such as FID and R-Precision do fail to fully validate our framework.
% 我们统计了每一轮的动作指标。从turn 0到turn 1后，FID从0.6443降到0.5326，Text-Motion Similarity 从 0.9257 提升到 0.9339；Turn 1/2 中 Local Preservation Ability 保持为 1.0000，说明局部编辑不会破坏未编辑区域。

% Our core goal is boosting controllability, editing accuracy and user satisfaction in iterative interaction instead of single-shot generation performance. 
% We conduct human evaluation of MIRO in terms of Semantic Alignment, Editing Accuracy and Overall Satisfaction. 
% 指标需要重新跑一下，现在暂时放到这里占一个位子
%结果显示，我们的方法在每一轮修改之后，可以做到blabla
% 充分说明我们方法的有效性
% \begin{table}[htbp]
% \centering
% \scriptsize
% \setlength{\tabcolsep}{2pt}
% \caption{Multi-turn performance results.}
% \label{tab:multiturn_perf}
% \begin{tabular}{p{0.13\linewidth}p{0.13\linewidth}p{0.22\linewidth}p{0.24\linewidth}p{0.22\linewidth}}
% \toprule
% \centering Turn &
% \centering FID$\downarrow$ &
% \centering Text-Motion\\ Similarity$\uparrow$ &
% \centering Local Modification\\ Success Rate$\uparrow$ &
% \centering Local Preservation\\ Ability$\uparrow$ \tabularnewline
% \midrule
% \centering Turn 0 & \centering 0.6443 & \centering 0.9257 & \centering -- & \centering -- \tabularnewline
% \centering Turn 1 & \centering 0.5326 & \centering 0.9339 & \centering 0.4045 & \centering 1.0000 \tabularnewline
% \centering Turn 2 & \centering 0.5326 & \centering 0.9306 & \centering 0.5000 & \centering 1.0000 \tabularnewline
% \bottomrule
% \end{tabular}
% \end{table}

\subsection{Ablation Study}
\label{sec: Ablation Studies}

\begin{table}[b]
\centering

\resizebox{\linewidth}{!}{
\begin{tabular}{ccccccc}
\toprule
\textbf{Cha.} & \textbf{Gen.} & \textbf{Che.} & \textbf{Edi.} &
\textbf{R-Prec@1$\uparrow$} &
\textbf{FID$\downarrow$} &
\textbf{Diversity$\rightarrow$} \\
\midrule

\ding{55} & \ding{51} & \ding{55} & \ding{55} &
0.485$\pm$.002 &
0.554$\pm$.015 &
9.557$\pm$.228 \\

\ding{55} & \ding{51} & \ding{51} & \ding{51} &
0.482$\pm$.004 &
0.578$\pm$.009 &
9.571$\pm$.234 \\

\ding{51} & \ding{51} & \ding{55} & \ding{51} &
0.483$\pm$.005 &
0.577$\pm$.010 &
9.571$\pm$.110 \\

\ding{51} & \ding{51} & \ding{51} & \ding{55} &
0.514$\pm$.005 &
0.531$\pm$.034 &
9.500$\pm$.522 \\

\midrule
\rowcolor{gray!10}
\ding{51} & \ding{51} & \ding{51} & \ding{51} &
\textbf{0.535$\pm$.003} &
\textbf{0.503$\pm$.009} &
\textbf{9.662$\pm$.065} \\
\bottomrule
\end{tabular}}
\caption{Ablation study of multi-agent components.}
\label{tab:component_ablation}
\end{table}
\subsubsection{Effect of Multi-agent Collaboration}
We ablated SMART by individually removing the Chatter, Checker, or Editor. As shown in Table \ref{tab:component_ablation}, each ablation degraded semantic alignment and motion quality, demonstrating the complementary contribution of each agent. Removing the Checker and Editor caused the performance drop, underscoring the importance of semantic feedback for accurate motion refinement. 

\subsubsection{Effect of Core Modules}
% To evaluate the effectiveness of the proposed Memory-based Self-Evolving Strategy.

\begin{figure}[!htbp]
\centering
\includegraphics[width=0.95\linewidth]{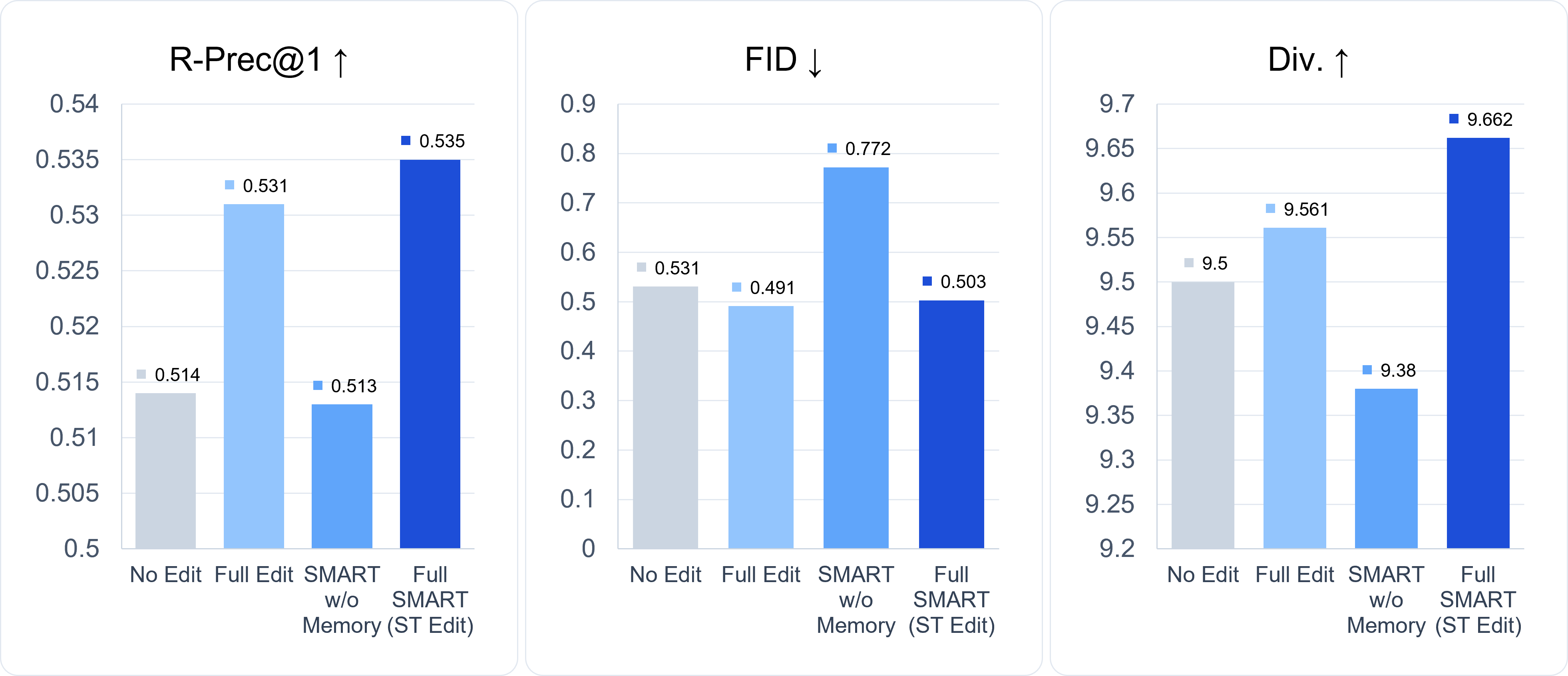}
\caption{
% Ablation study of the core modules in SMART.
Ablation results of memory-based refinement and editing strategies.
}
\label{fig:Edit_ablation}
\end{figure}

We evaluated the memory-based self-evolving module and spatio-temporal (ST) editing through ablation studies. As shown in Fig.~\ref{fig:Edit_ablation}, full SMART achieved the highest R-Prec@1 and diversity. Compared with full editing, ST editing improved R-Prec@1 from 0.531 to 0.535 and diversity from 9.561 to 9.662, with a slightly higher FID of 0.503 versus 0.491. Removing memory reduced R-Prec@1 and diversity while substantially increasing FID, demonstrating the benefit of experience reuse for motion refinement.

% both editing strategies improve semantic alignment over the No Edit baseline, demonstrating the necessity of motion refinement. Compared with Full Edit, our ST Edit further improves R-Precision from 0.531 to 0.535 and Diversity from 9.561 to 9.662, showing that localized spatio-temporal refinement better preserves correct motion regions while maintaining high generation quality. 

\begin{figure}[!ht]
\centering
\includegraphics[width=\linewidth]{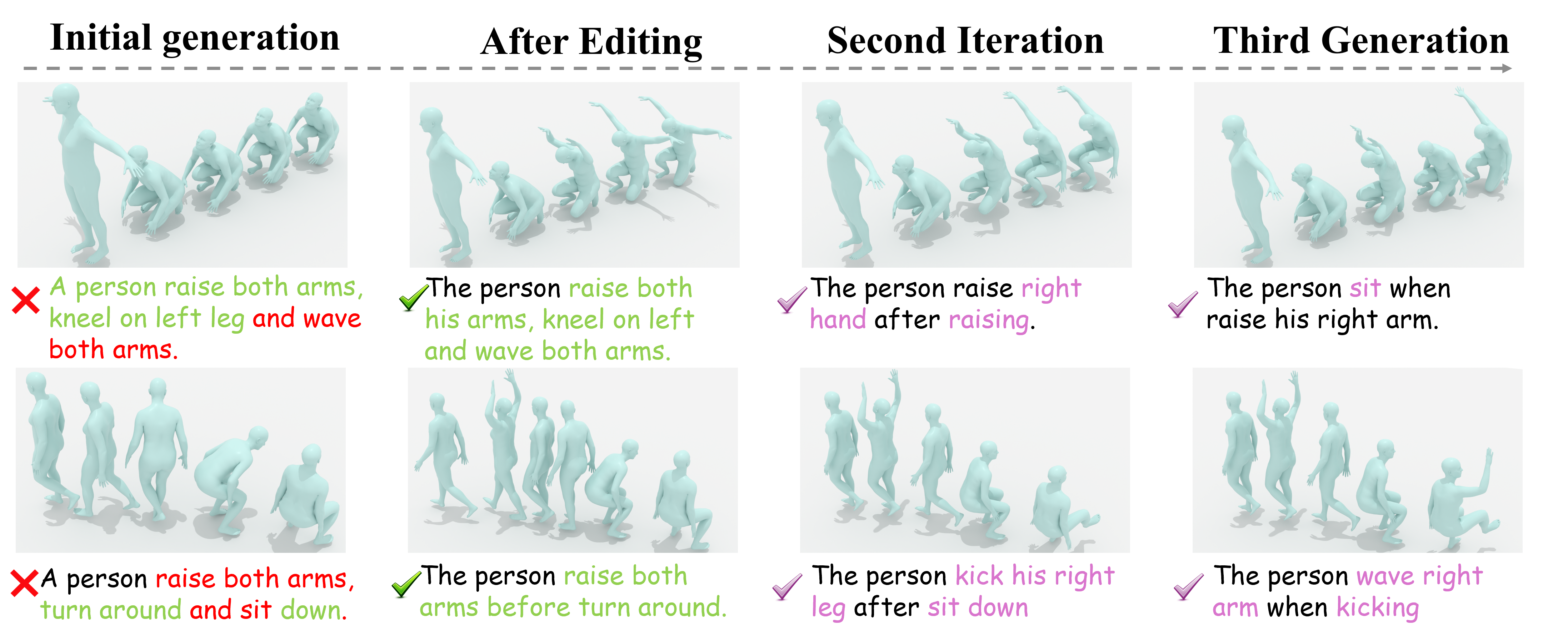}
\caption{
Illustrative example of the iterative motion refinement process, with semantic misalignment in red, semantic alignment in green, and personality instruction in purple.
%Starting from an initial motion, the system progressively improves motion quality through feedback-driven updates. Finally, user-provided editing instructions are incorporated to further refine the motion, enabling controllable and incremental motion generation.
}
\label{fig:self-evolving}
\end{figure}

\subsection{Human Evaluation}

\subsubsection{Pair-wise Preference Study}
\begin{figure}[!htbp]
\centering

\begin{subfigure}[c]{0.46\linewidth}
    \centering
    \includegraphics[height=3.1cm]{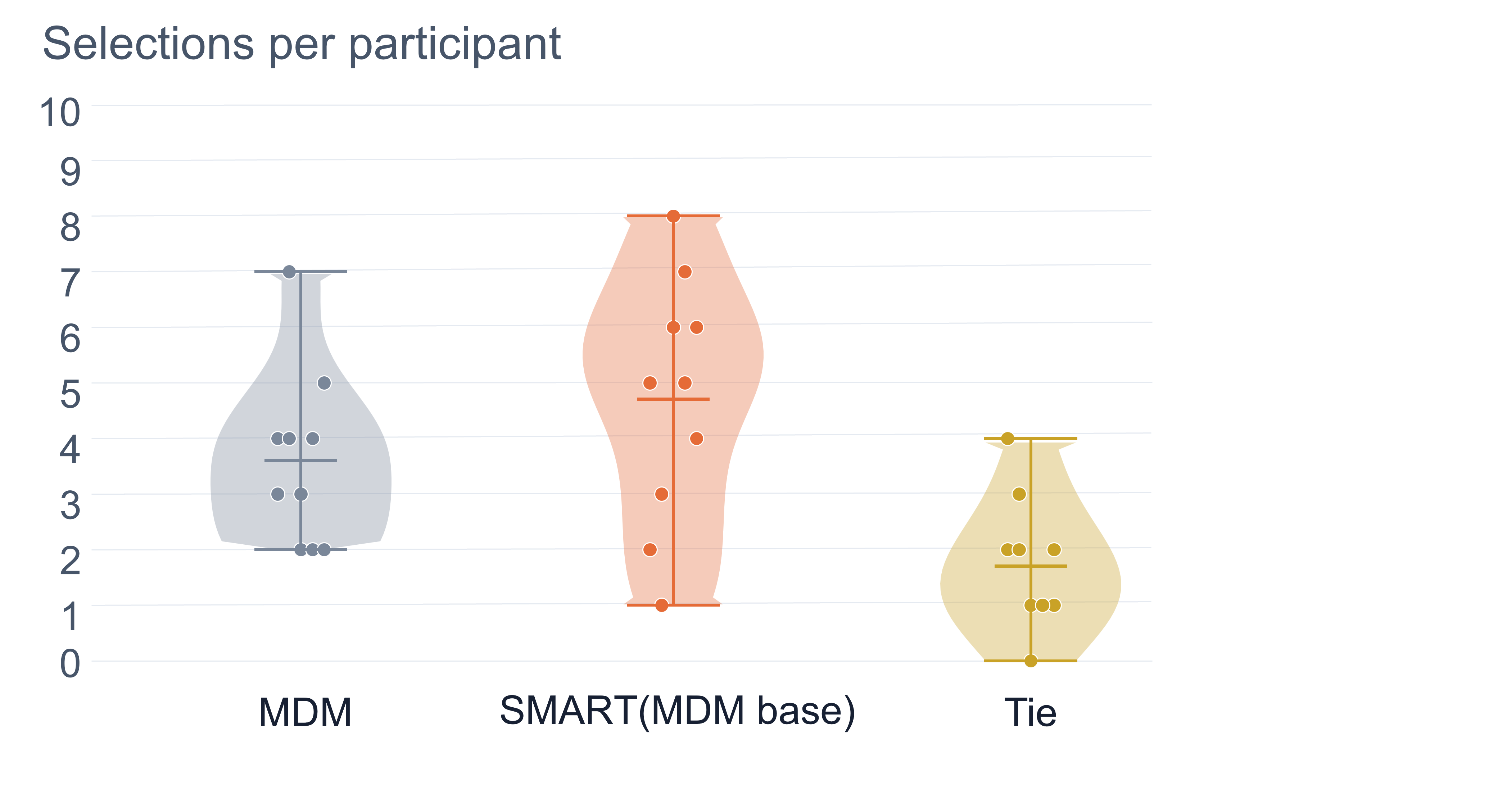}
    \caption{Pair-wise preference study.}
    \label{fig:violin}
\end{subfigure}
\hfill
\begin{subfigure}[c]{0.46\linewidth}
    \centering
    \includegraphics[height=3.1cm]{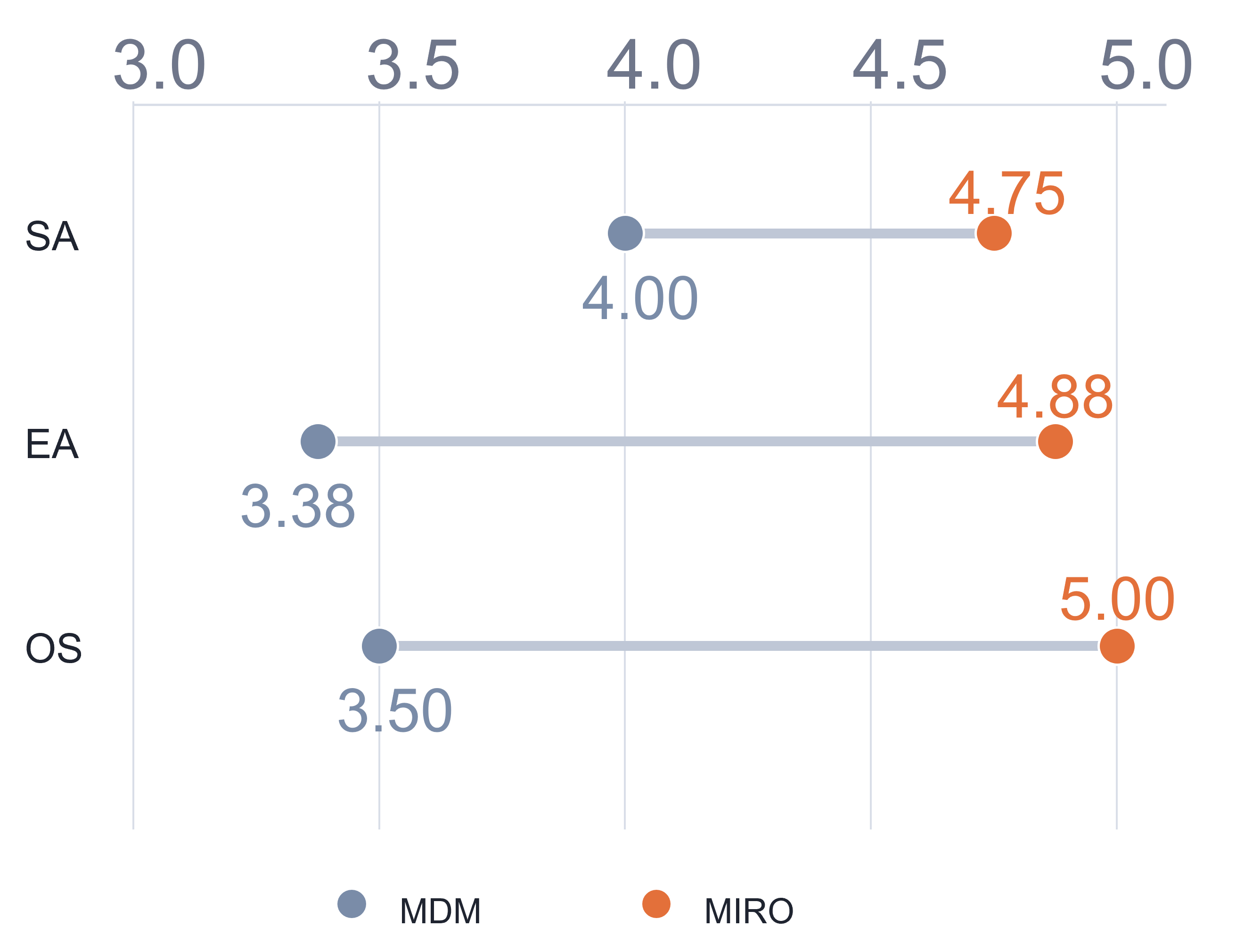}
    \caption{User evaluation results.}
    \label{fig:dumbbell}
\end{subfigure}

\caption{Human evaluation results of MDM and SMART.}
\label{fig:human_evaluation}
\end{figure}

We randomly sampled 200 text descriptions from the HumanML3D test set and generated one motion with MDM and one with SMART for each description. The resulting pairs are presented anonymously in random order. Each of the 20 participants evaluates 10 pairs based on semantic consistency and motion naturalness, yielding 200 evaluations. As shown in Fig.~\ref{fig:violin}, SMART receives more preferences than MDM, indicating improved semantic alignment and perceptual quality.
\subsubsection{Human–VLM Preference Consistency}
\begin{figure}[!htbp]
\centering

\begin{subfigure}[ht]{0.48\linewidth}
    \centering
    \includegraphics[
        width=\linewidth,
        height=3.2cm,
        keepaspectratio
    ]{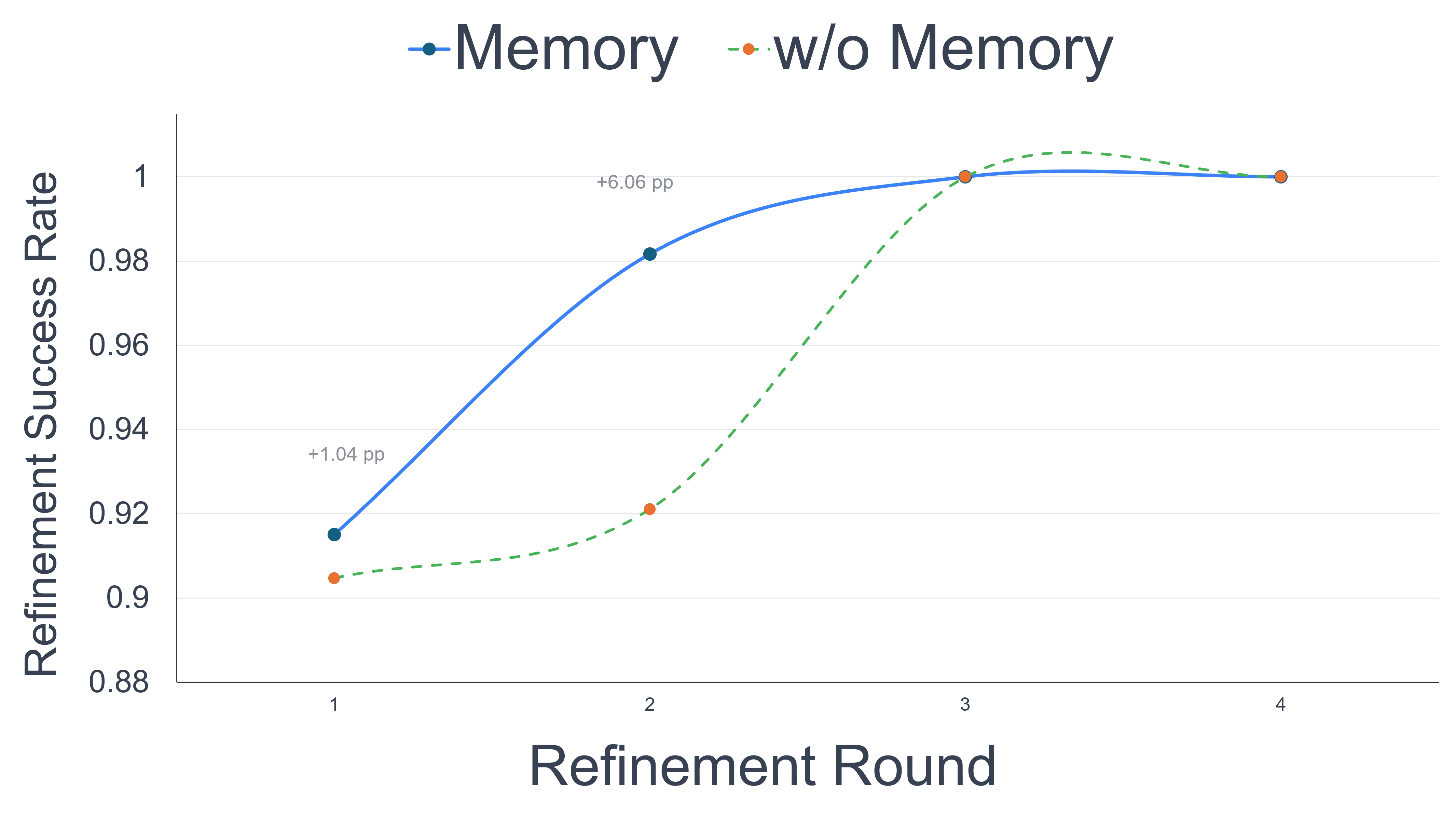}
    \caption{
    % Memory-based refinement results.
    Performance across iterative refinement steps.
    }
    \label{fig:memory}
\end{subfigure}
\hfill
\begin{subfigure}[ht]{0.48\linewidth}
    \centering
    \includegraphics[
        width=\linewidth,
        height=3.2cm,
        keepaspectratio
    ]{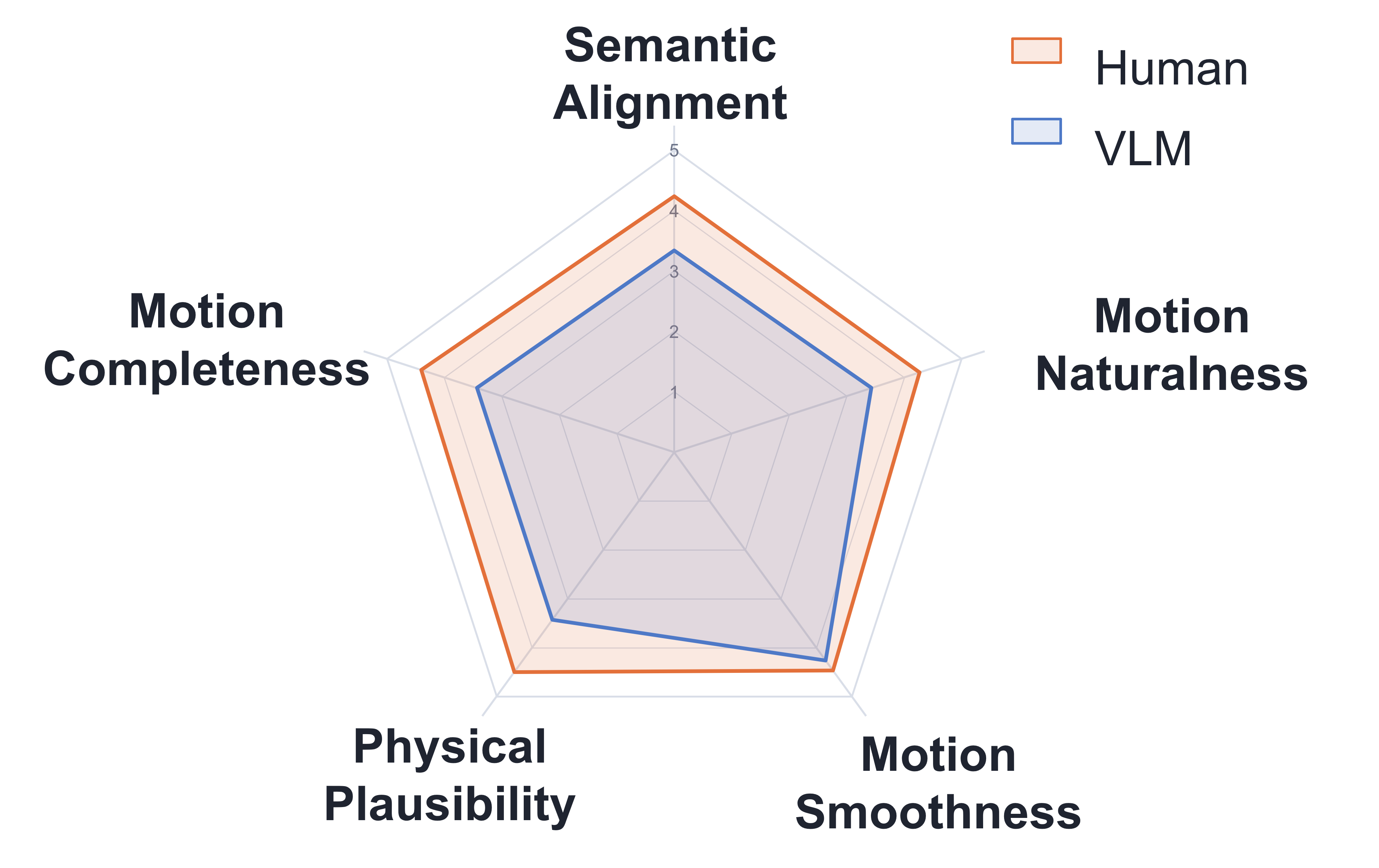}
    \caption{Human-VLM evaluation comparison.}
    \label{fig:radar}
\end{subfigure}

\caption{Evaluation of memory-based refinement and VLM-based motion assessment.}
\label{fig:memory_vlm_evaluation}
\end{figure}

We randomly sample 200 motion pairs from HumanML3D and evaluate them using five criteria as the human study: Semantic Alignment, Motion Naturalness, Motion Coherence, Physical Plausibility, and Action Completeness. Each criterion is rated on a five-point Likert scale. As shown in Fig.~\ref{fig:radar}, VLM and human evaluations exhibit broadly similar trends, supporting VLM assessment as a complementary reference for motion evaluation.

% \begin{figure}[htbp]
% \centering
% \begin{subfigure}[c]{0.48\linewidth}
%     \centering
%     \includegraphics[height=3.5cm]{image/violin.png}
%     \caption{Pair-wise preference study.}
%     \label{fig:violin}
% \end{subfigure}
% \hfill
% \begin{subfigure}[c]{0.48\linewidth}
%     \centering
%     \includegraphics[height=3.5cm]{image/dumbbell.png}
%     \caption{User evaluation results.}
%     \label{fig:dumbbell}
% \end{subfigure}

% \caption{Human evaluation results of MDM and MIRO.}
% \label{fig:human_evaluation}
% \end{figure}

% \caption{Human evaluation results of MDM and MIRO.}
% \label{fig:human_evaluation}
% \end{figure}
\subsubsection{User Evaluation}

% To evaluate the practical usability of MIRO, we conduct a human study with 12 human participants. Participants were asked to evaluate the generated motions from three aspects: Semantic Alignment, Editing Accuracy, and Overall Satisfaction. All scores range from 1 to 5.

% \begin{table}[htbp]
% \centering
% \caption{Human evaluation results.}
% \label{tab:human_eval}
% \resizebox{\linewidth}{!}{
% \begin{tabular}{lccc}
% \toprule
% \textbf{Method} &
% \textbf{Semantic Alignment$\uparrow$} &
% \textbf{Editing Accuracy$\uparrow$} &
% \textbf{Overall Satisfaction$\uparrow$} \\
% \midrule
% Baseline (MDM) &
% 4.08 &
% 3.67 &
% 3.67 \\
% \midrule
% \rowcolor{gray!10}
% \textbf{MIRO} &
% \textbf{4.58} &
% \textbf{4.75} &
% \textbf{4.75} \\
% \bottomrule
% \end{tabular}}
% \end{table}
% As shown in Fig.~\ref{fig:dumbbell}, MIRO consistently outperforms the baseline across all three evaluation metrics. The higher Semantic Alignment score indicates that multi-turn interaction better captures user intentions, while the improvement in Editing Accuracy demonstrates the effectiveness of the proposed semantic-guided motion editing mechanism. 
% Moreover, the highest Overall Satisfaction score confirms that MIRO provides a more natural and effective conversational motion editing experience.

We conducted a user study with 12 participants to assess the practical usability of SMART. Participants evaluate generated motions in terms of Semantic Alignment, Editing Accuracy, and Overall Satisfaction on a five-point scale. As shown in Fig.~\ref{fig:dumbbell}, SMART outperforms the baseline across all three metrics, indicating improved intent understanding, effective local editing, and a more natural conversational motion editing experience.

\begin{figure}[h]
\centering
\includegraphics[width=\linewidth]{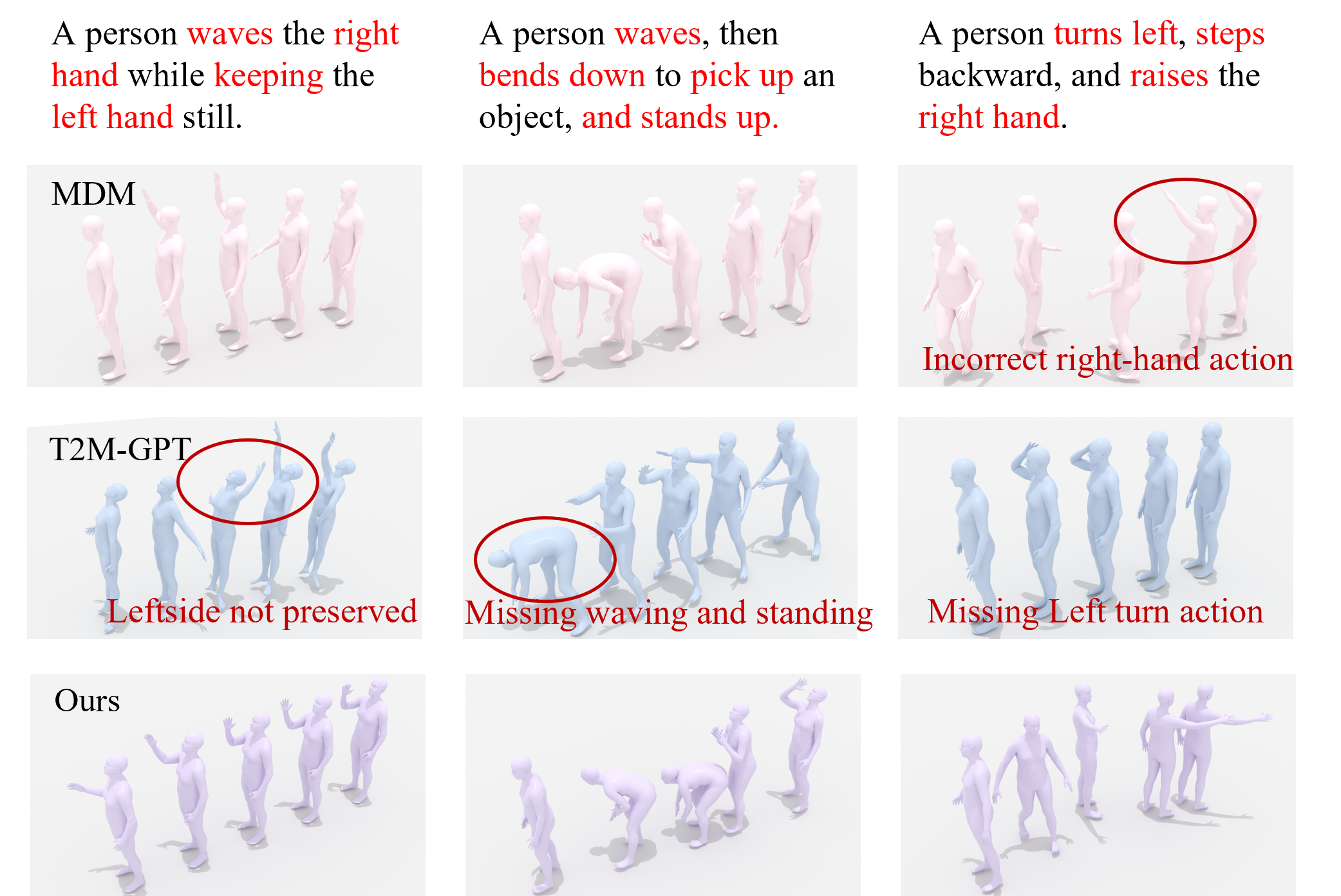}
\caption{
Qualitative comparison of different text-to-motion methods on compositional prompts.
Key action phrases are highlighted in red.
}
\label{fig:vis_com}
\end{figure}

\subsection{Visualization}
% \subsubsection{Visualization}
% We present visual comparison results to further demonstrate the effectiveness of SMART. As shown in 
Fig.~\ref{fig:vis_com} compares MDM, T2M-GPT, and SMART on compositional prompts.
SMART better captures multiple actions, action order, and fine-grained body-part constraints. 
% generates motions that better align with the input text, particularly for compositional instructions involving multiple actions and fine-grained body movements. Moreover, our method produces more natural and coherent motion sequences while preserving semantic consistency throughout the generation process. 
% And Fig.~\ref{fig:self-evolving} further illustrates the iterative refinement process of SMART. Starting from an initial motion with semantic mismatches, SMART progressively refines only the problematic regions according to user feedback while preserving previously correct motion segments. After several interaction rounds, the generated motion gradually converges to the target instruction, demonstrating the effectiveness of the proposed localized conversational refinement framework.
Figure~\ref{fig:self-evolving} further illustrates the iterative refinement process, where semantic mismatches are progressively corrected according to user feedback while previously accepted motion content is largely preserved. These results
qualitatively demonstrate the effectiveness of SMART in compositional generation and localized conversational refinement.

\section{Conclusion}
% We have proposed an interactive motion generation framework that integrates structured intent parsing, modular function orchestration, and multi-round semantic optimization. 
% % By decoupling generation and editing into callable motion functions and enabling feedback-driven semantic routing, the system supports flexible and context-aware motion synthesis.

% The introduced \textbf{DeltaEditor}, trained under joint edit supervision and delta supervision, demonstrates strong semantic correction capabilities, particularly in handling missing actions, temporal misalignment, and attribute inconsistencies.

% Extensive experiments on HumanML3D, KIT-ML, and MotionFix datasets show that our framework achieves competitive generation quality while significantly improving semantic fidelity through efficient and non-destructive editing. Ablation studies further validate the necessity of modular design, chained planning, semantic verification, and dual supervision.

% This work underscores the importance of interactive optimization in text-driven motion generation systems and establishes a scalable foundation for future extensions, including style adaptation, multi-agent collaboration, and real-time user-in-the-loop generation.
In this paper, we introduced the \textbf{S}elf-evolving \textbf{M}ulti-\textbf{A}gent \textbf{R}efinemen\textbf{T} Framework (SMART) for conversational text-to-motion generation. By integrating the Chatter, Generator, Editor, and Checker agents, SMART demonstrates 
% significant improvements
consistent improvements
in motion quality and semantic consistency. Our ablation studies highlight the crucial contributions of the four agents, with the full SMART configuration yielding superior performance over individual agents. Additionally, our evaluation of editing strategies showed that Spatio-Temporal Editing outperforms other methods, ensuring high-quality motions while maintaining temporal coherence.

\bibliography{ref}


\end{document}